\documentclass[lettersize,journal]{IEEEtran}
\usepackage{amsmath,amsfonts}
\usepackage{algorithmic}
\usepackage{algorithm}
\usepackage{array}
\usepackage[caption=false,font=normalsize,labelfont=sf,textfont=sf]{subfig}
\usepackage{textcomp}
\usepackage{booktabs}
\usepackage{stfloats}
\usepackage{url}
\usepackage{verbatim}
\usepackage{graphicx}
\usepackage{cite}
\usepackage{tabularx}
\usepackage{algorithm}  
\usepackage{algorithmic} 
\usepackage{hyperref}
\hypersetup{hidelinks}
\usepackage{tcolorbox}
\tcbuselibrary{skins, breakable, listings}
\usepackage{xcolor}
\usepackage{enumitem}
\usepackage{booktabs}

\colorlet{hdrtcol}{blue!45!black}
\colorlet{mainbg}{blue!4!white}
\colorlet{avboxbg}{green!7!white}
\colorlet{hdvboxbg}{orange!9!white}
\colorlet{instbg}{gray!4!white}

\begin{document}

\title{SVBRD-LLM: Self-Verifying Behavioral Rule Discovery for Autonomous Vehicle Identification}

\author{Xiangyu Li$^{1}$, Tianyi Wang$^{1}$, Junfeng Jiao$^{2}$, Christian Claudel$^{1}$,  and Zhaomiao Guo$^{1\dag}$

\thanks{$^{\dag}$Corresponding author: Zhaomiao Guo.}%
\thanks{$^{1}$Fariborz Maseeh Department of Civil, Architectural, and Environmental Engineering, The University of Texas at Austin, Austin, TX 78712, USA.
{\tt\small \{xiangyu\_li, bonny.wang, christian.claudel, zguo\}@utexas.edu}}%
\thanks{$^{2}$School of Architecture, The University of Texas at Austin, Austin, TX 78712, USA.
 	{\tt\small jjiao@austin.utexas.edu}}%

}





\maketitle

\begin{abstract}
As autonomous vehicles (AVs) become increasingly deployed on public roads, understanding their real-world behavior characteristics is critical for traffic safety analysis, regulatory oversight, and public acceptance. 
However, existing data-driven approaches often lack interpretability and fail to provide verifiable explanations of AV driving behavior in mixed traffic environments.
To address this challenge, this paper proposes SVBRD-LLM, a self-verifying behavioral rule discovery framework that automatically extracts interpretable behavioral rules from real-world traffic videos through zero-shot large language model (LLM) reasoning. 
The framework first extracts vehicle trajectories from video data using YOLOv26-based detection and ByteTrack-based tracking, followed by the computation of kinematic features and contextual information.
It then employs GPT-5 zero-shot prompting to perform comparative behavioral analysis between AVs and human-driven vehicles (HDVs) across lane-changing and normal driving behaviors, generating 26 structured rule hypotheses that comprises both numerical thresholds and statistical behavioral patterns.
These rules are subsequently evaluated through the AV identification task using an independent validation dataset, and iteratively refined through failure case analysis to filter spurious correlations and improve robustness.
The resulting rule library contains 20 high-confidence behavioral rules, each including semantic description, quantitative thresholds or behavioral patterns, applicable context, and validation confidence.
Experiments conducted on over 1,500 hours of real-world traffic videos from Waymo's commercial operating area demonstrate that the proposed framework achieves 90.0\% accuracy and 93.3\% F1-score in AV identification, with 98.0\% recall. 
The discovered rules reveal distinctive characteristics of AVs in motion smoothness, conservative behavior, and lane discipline, providing valuable insights for traffic safety assessment, regulatory compliance, and adaptive traffic management during the transition to mixed traffic environment. The dataset used in this study is available at: \href{https://huggingface.co/datasets/Ryan-xiangyu-ut/svbrd-llm-roadside-video-av/tree/main}{svbrd-llm-roadside-video-av}.
\end{abstract}


\section{Introduction}

\IEEEPARstart{T}{he} rapid development of autonomous driving technology is reshaping modern transportation systems. 
Meanwhile, the growing deployment of robotaxi fleets, including those operated by Waymo, Tesla and Zoox, has accelerated the integration of Autonomous Vehicles (AVs) into ordinary urban traffic, forming mixed traffic flow \cite{zhao2025robotaxis}. 
In many cases, AVs no longer present strongly distinguishable external appearances relative to Human-Driven Vehicles (HDVs), thereby increasing the difficulty of recognizing AVs based solely on visual cues and underscoring the need to identify them through behavioral characteristics \cite{li2026characteristics}. 
Existing research shows that AV penetration rate can significantly influence traffic dynamics in mixed traffic environments, for example, affecting conflict rate \cite{b1}, Time-To-Collision (TTC) metrics \cite{wang2025hlcg}, and merging efficiency \cite{wang2024research} in mixed traffic flow. 
At the microscopic level, evidence indicates that human drivers adjust their driving strategies when following AVs, resulting in observable changes in time headway, acceleration patterns, and lane-changing behavior \cite{b2,b3}. 
These behavioral differences not only affect the stability and safety of local traffic flow but may also impact the broader traffic system through traffic wave propagation. 
Therefore, understanding and identifying the behavioral characteristics of AVs is crucial for traffic safety assessment, intelligent traffic management, regulatory policy development, and enhancing public acceptance of autonomous driving technologies.

However, existing research has two major limitations in understanding AV behavior. 
First, mainstream studies focus on trajectory prediction tasks, which estimate future positions given historical trajectory data \cite{wang2026kept,b5}. 
While these approaches can achieve high predictive accuracy, they rarely provide systematic explanations for the underlying behavioral mechanisms, such as why AVs choose to decelerate earlier in specific traffic scenarios or how AV lane-changing strategies differ fundamentally from those of human drivers. 
Second, many deep learning-based vehicle behavior recognition methods achieve excellent classification performance on simulation-generated data, yet their decision processes remain unexplainable \cite{b6}. 
Traditional explainability methods, such as post-hoc analysis techniques using SHAP value analysis \cite{b7} or visualization of attention mechanisms \cite{b8}, can reveal correlations between input features and model outputs.
However, they fails to generate verifiable and generalizable behavioral rules across different traffic environments.
This lack of interpretability poses a serious problem in safety-critical transportation domains, where regulators, insurance companies, and the public need to understand the decision logic of AVs \cite{b9}.

\begin{figure*}[htbp]
\centerline{\includegraphics[width=0.75\linewidth]{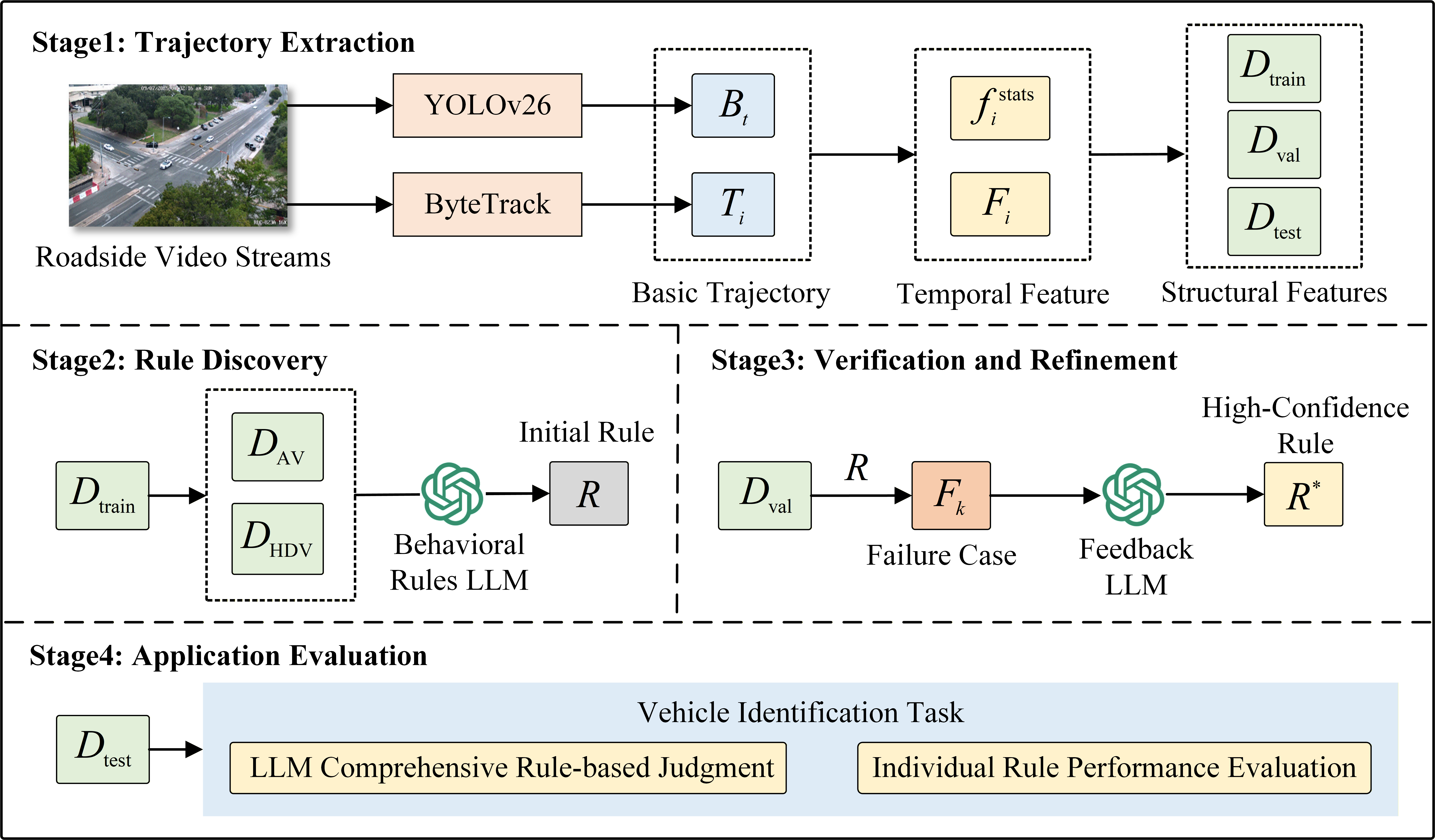}}
\caption{Overall architecture of the proposed SVBRD-LLM framework showing the three-stage pipeline: rule discovery from labeled trajectories, verification and refinement through iterative testing, and application evaluation on multiple downstream tasks.}
\label{fig:framework}
\end{figure*}

In recent years, Large Language Models (LLMs) have been increasingly introduced into the transportation domain, providing new techniques for traffic behavior understanding and analysis \cite{zhang2025ccma,li2026llm}. 
By modeling sequential patterns through natural language representations, LLM-based approaches can transform trajectory modeling problems into sequence prediction tasks by discretizing continuous trajectories into tokens \cite{b16,b17,b18,b19,b20}. 
Furthermore, LLMs have also been integrated into traffic operations and control, enabling applications such as lane-change prediction, end-to-end autonomous driving decision-making, and signal control \cite{b21,b22,b23,b24,b25,b32}. 
Meanwhile, motion pattern-based studies show that human drivers adjust their car-following and lane-changing behavior when interacting with AVs \cite{wang2025impact}.
Motion-related variables can effectively indicate driving style differences \cite{b30}, and motion pattern classifiers can reliably distinguish AVs from HDVs based on trajectory features \cite{b33,b34}, providing a promising foundation for behavior-based vehicle identification.

However, the application of LLMs to traffic behavior analysis remains limited by two key challenges. 
On one hand, fine-tuning for specific tasks, while capable of improving accuracy, often leads to loss of in-context learning ability.
As a result, models may overfit to specific task input-output formats and lose flexibility when encountering new scenarios \cite{b12,wang2025rad,cui2025board,zhao2025vlm}. 
On the other hand, research on LLM reasoning reliability shows that models often struggle to self-correct reasoning errors without external verification mechanisms \cite{b26,b27,b28,b29}. 
In contrast, zero-shot prompt engineering preserves the model's original reasoning capabilities while avoiding task-specific retraining \cite{zhang2024semantic,choi2026texting}.
Zero-shot prompts presented in natural language form are inherently interpretable, require minimal labeled data, and maintain the broad knowledge and reasoning abilities learned during pre-training \cite{b13}. 
Prior research shows that for tasks requiring abstract reasoning and pattern discovery, in-context learning outperforms fine-tuning methods \cite{b14}.
Moreover, prompt-based methods can generate more transparent and interpretable decision processes through explicit reasoning chains  \cite{b15}. 
Nevertheless, existing work rarely integrates explicit and verifiable behavioral rules with behavior-based AV identification in mixed traffic environment within a unified analytical framework.

To address these challenges, this paper proposes \textbf{SVBRD-LLM} (Self-Verifying Behavioral Rule Discovery via Large Language Model), a unified framework which leverages the zero-shot reasoning capabilities of LLMs to automatically discover, verify, and apply interpretable behavioral rules from real-world traffic videos featuring commercial AVs (Figure \ref{fig:framework}). 
Specifically, the framework transforms behavioral rule discovery into a comparative analysis and pattern induction problem, enabling the LLM to to extract structured knowledge through natural language prompts. 
The overall workflow consists of three stages. 
First, the rule discovery stage generates structured rule hypotheses comprising numerical rules based on quantitative thresholds and behavioral statistical rules based on behavioral patterns through comparative analysis of labeled AV and HDV trajectory data.
Second, tests rule performance on the vehicle identification task using an independent validation dataset, performs reflective analysis of failure cases, and iteratively refines the rules.
Finally, the evaluation stage examines the generalization capability of the discovered rule set on the AV identification downstream task using an independent test dataset.
The main contributions of this paper are summarized as follows:

\begin{itemize}
\item We propose a novel AV behavior analysis framework based on zero-shot prompt engineering that extracts an interpretable behavioral rule knowledge base from real-world trajectory data, avoiding the loss of generalization ability caused by task-specific fine-tuning while maintaining strong reasoning ability. 
\item We propose a failure-case reflection-based rule refinement mechanism that guides the model to analyze prediction error causes for the vehicle identification task through iterative prompting, outputs refinement suggestions in natural language form, filters unreliable rules, and improves rule applicability conditions and thresholds.
\item We construct a verified rule library containing 20 high-confidence rules derived from real-world traffic videos in Waymo's commercial operational area, achieving high-accuracy AV identification while providing interpretable and verifiable behavioral pattern insights for traffic safety assessment, adaptive traffic management, and regulatory policy development.
\end{itemize}

The remainder of this paper is organized as follows: Section \ref{sec:2} describes the methodology of the SVBRD-LLM framework; Section \ref{sec:3} presents the experimental setup and results; Section \ref{sec:4} concludes the paper and discusses future work.

\section{Methodology}
\label{sec:2}

This section details the complete methodology of the SVBRD-LLM framework. The overall algorithmic workflow of the framework is shown in Figure~\ref{fig:framework}.

\subsection{Problem Formulation}

Given a set of traffic videos $\mathcal{V} = \{V_1, V_2, \ldots, V_N\}$ and a small number of labeled AV and HDV samples $\mathcal{D}_{\text{labeled}} = \{(\tau_i, y_i)\}_{i=1}^M$, where $\tau_i$ is the trajectory of vehicle $i$ and $y_i \in \{\text{AV}, \text{HDV}\}$ is the label, our objectives include three levels.

First, automatically extract a set of behavioral rules that distinguish AVs from HDVs:
\begin{equation}
\mathcal{R} = \mathcal{R}_{\text{numerical}} \cup \mathcal{R}_{\text{behavioral}},
\end{equation}
where the rules are divided into two categories:
\begin{itemize}
\item \textbf{Numerical Rules} $\mathcal{R}_{\text{numerical}}$: Decision rules based on quantified features.
\item \textbf{Behavioral Statistical Rules} $\mathcal{R}_{\text{behavioral}}$: Decision rules based on behavioral pattern statistics.
\end{itemize}

Each rule can be represented as a triple:
\begin{equation}
r_k = \langle d_k, \phi_k, c_k \rangle,
\end{equation}
where $d_k$ is the natural language description of the rule, $\phi_k: \mathcal{T} \rightarrow \{0, 1\}$ is a decision function based on trajectory features $\mathcal{T}$, and $c_k$ represents the contextual constraints for applicability.

Second, evaluate the effectiveness of each rule through prediction tasks and filter unreliable rules. The confidence of rule $r_k$ on validation set $\mathcal{D}_{\text{val}}$ is defined as:
\begin{equation}
\text{Confidence}(r_k) = \frac{1}{|\mathcal{D}_{\text{val}}|} \sum_{(\tau_i, y_i) \in \mathcal{D}_{\text{val}}} \mathbb{1}[\phi_k(\tau_i) = y_i].
\end{equation}
The set of verified rules is defined as:
\begin{equation}
\mathcal{R}^* = \{r_k \in \mathcal{R} \mid \text{Confidence}(r_k) \geq \theta\},
\end{equation}
where $\theta$ is a preset confidence threshold.

Finally, determine vehicle types for unlabeled trajectories based on the verified rules. 
Unlike traditional end-to-end classification methods, this framework emphasizes the interpretability and verifiability of rules, requiring that each identification decision is supported by clear behavioral evidence.

\subsection{Trajectory Extraction and Scene Understanding}

\subsubsection{Basic Trajectory Extraction}

We employ YOLOv26 \cite{sapkota2026yoloe} as the object detector to detect all vehicles in each frame $V_t$, outputting a set of bounding boxes:
\begin{equation}
\mathcal{B}_t = \{b_t^j = (x_j, y_j, w_j, h_j, s_j)\}_{j=1}^{N_t},
\end{equation}
where $(x_j, y_j)$ is the bounding box center coordinates, $(w_j, h_j)$ is the width and height, and $s_j$ is the detection confidence score.

\begin{figure}[htbp]
\centerline{\includegraphics[width=\columnwidth]{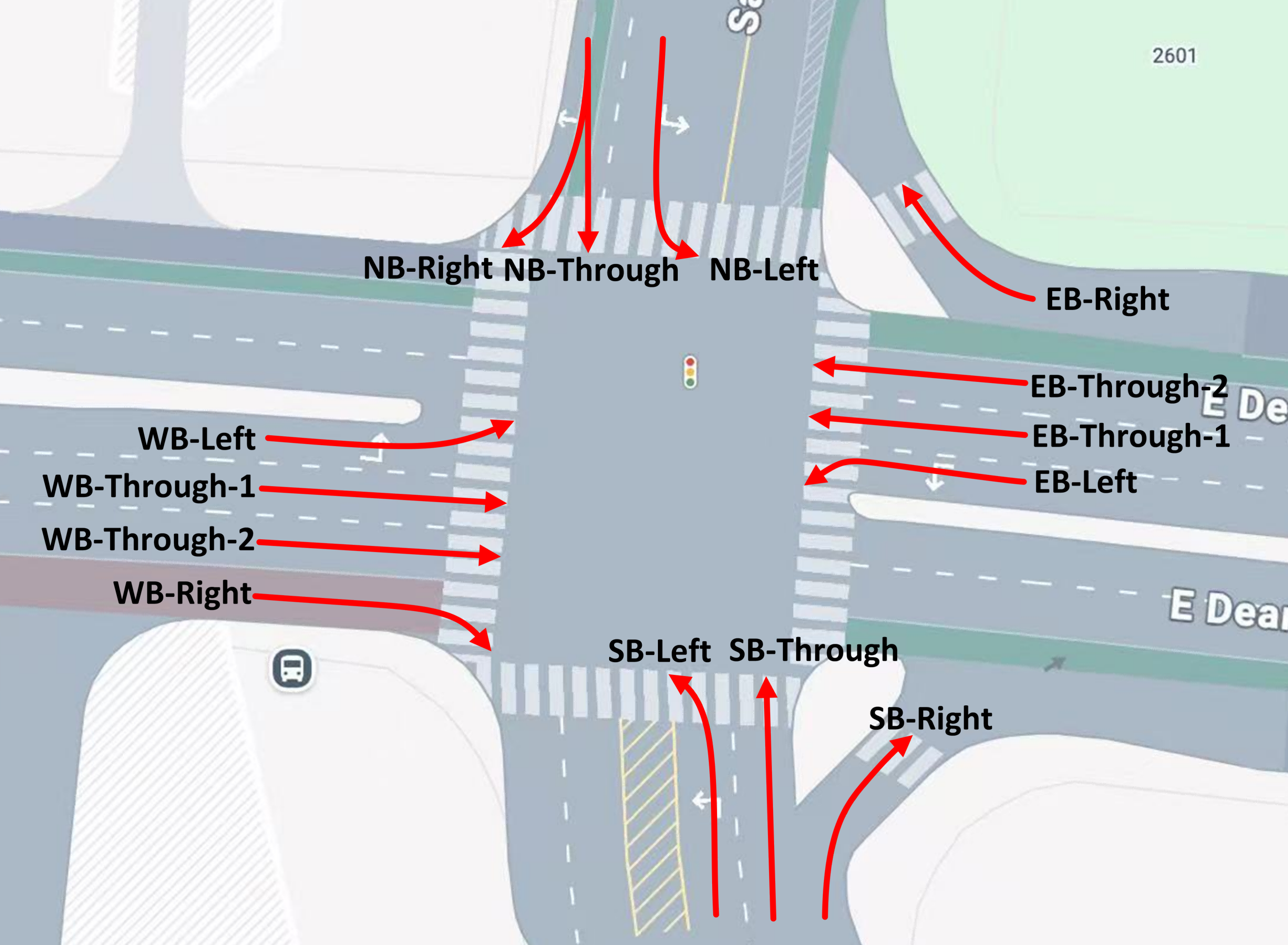}}
\caption{Lane identification and naming scheme for the signalized intersection. Each approach is labeled by geographic orientation (Eastbound, Westbound, Northbound, Southbound) and further divided into functional lanes (Left, Through, Right).}
\label{fig:lane_definition}
\end{figure}

The ByteTrack multi-object tracking algorithm \cite{zhang2022bytetrack} is used to associate detection results across the temporal dimension, assigning a unique ID $i$ to each vehicle and constructing a complete trajectory:
\begin{equation}
\tau_i = \{(u_i^t, v_i^t)\}_{t=t_{\text{start}}^i}^{t_{\text{end}}^i},
\end{equation}
where $(u_i^t, v_i^t)$ is image pixel coordinates.

The preliminary trajectories obtained from tracking undergo Kalman filter \cite{khodarahmi2023review} smoothing and removal of stationary trajectories with excessively low average pixel velocities.

\subsubsection{Spatial Calibration and Physical Coordinate Transformation}

To convert pixel coordinates to real-world physical coordinates, we select $n$ reference point pairs in the scene, each containing image pixel coordinates $(u_j, v_j)$ and corresponding real-world coordinates $(X_j, Y_j)$ (in meters). 
Through these reference point pairs, we solve for the homography matrix $\mathbf{H} \in \mathbb{R}^{3 \times 3}$ to establish a perspective transformation relationship:
\begin{equation}
\begin{bmatrix} u \\ v \\ 1 \end{bmatrix} \sim \mathbf{H} \begin{bmatrix} X \\ Y \\ 1 \end{bmatrix},
\end{equation}
where $\sim$ denotes equivalence under homogeneous coordinates. 
The homography matrix is solved using least squares:
\begin{equation}
\min_{\mathbf{H}} \sum_{j=1}^{n} \left\| \begin{bmatrix} u_j \\ v_j \\ 1 \end{bmatrix} - \lambda_j \mathbf{H} \begin{bmatrix} X_j \\ Y_j \\ 1 \end{bmatrix} \right\|^2.
\end{equation}
For any pixel point $(u_i^t, v_i^t)$, we convert to world coordinates using $\mathbf{H}^{-1}$:
\begin{equation}
\begin{bmatrix} X_i^t \\ Y_i^t \\ z_i^t \end{bmatrix} = \mathbf{H}^{-1} \begin{bmatrix} u_i^t \\ v_i^t \\ 1 \end{bmatrix}.
\end{equation}
After normalization, we obtain the real physical coordinates:
\begin{equation}
X_i^t = \frac{X_i^t}{z_i^t}, \quad Y_i^t = \frac{Y_i^t}{z_i^t}.
\end{equation}

\subsubsection{Lane Identification and Naming}

Based on the geometric structure and orientation of the scene, we define lane regions for each approach of the intersection. As illustrated in Figure~\ref{fig:lane_definition}, each approach is named according to its geographic orientation, for example:

\begin{itemize}
\item \textbf{Eastbound (EB)}: EB-Left, EB-Through-1, EB-Through-2, EB-Right.
\item \textbf{Westbound (WB)}: WB-Left, WB-Through-1, WB-Through-2, WB-Right.
\item \textbf{Northbound (NB)}: NB-Left, NB-Through, NB-Right.
\item \textbf{Southbound (SB)}: SB-Left, SB-Through, SB-Right.
\end{itemize}

For each trajectory point $(X_i^t, Y_i^t)$, we determine its lane $l_i^t$ through spatial geometric judgment. Lane changes are identified by detecting changes in lane labels across consecutive time steps:
\begin{itemize}
\item \textbf{Lane Change}: $l_i^t \neq l_i^{t+1}$ and both lanes belong to the same approach.
\item \textbf{Turn}: A vehicle exits one approach and enters another.
\end{itemize}

\subsubsection{Traffic Signal State Inference}

Since the camera view makes it difficult to directly identify signal light colors, we infer signal states through vehicle behavior. First, we define the stop line position $S_{\text{approach}}$ for each approach. At time $t$, for a given approach, we identify the vehicle $i^*$ closest to the stop line and analyze its behavior:

\begin{itemize}
\item \textbf{Red Light Inference}: If $d(i^*, S) < d_{\text{threshold}}$, $v_{i^*}^t \approx 0$, and the stopping duration exceeds threshold $T_{\text{stop}}$, we infer the signal is red for that direction.
\item \textbf{Green Light Inference}: If the lead vehicle's speed $v_{i^*}^t > v_{\text{threshold}}$ and the vehicle passes through the stop line, we infer the signal is green for that direction.
\item \textbf{Yellow Light Inference}: If the lead vehicle continuously decelerates but ultimately passes through the stop line, we infer the signal may be yellow.
\end{itemize}

Within time window $[t-\Delta T, t]$, we aggregate behaviors of multiple lead vehicles through voting to determine the most likely signal state. 
We annotate signal context for each vehicle at each time step:
\begin{equation}
\text{Signal}_i^t \in \{\text{Red}, \text{Green}, \text{Yellow}, \text{Unknown}\}.
\end{equation}

\subsubsection{Kinematic Feature Computation}

For each valid trajectory $\tau_i$, we use the central difference method to calculate instantaneous velocity and acceleration based on physical coordinates:
\begin{equation}
v_i^t = \frac{\sqrt{(X_i^{t+1} - X_i^{t-1})^2 + (Y_i^{t+1} - Y_i^{t-1})^2}}{2\Delta t},
\end{equation}
\begin{equation}
a_i^t = \frac{v_i^{t+1} - v_i^{t-1}}{2\Delta t},
\end{equation}
where $\Delta t = 1/f$ is the time interval and $f$ is the video frame rate. 
Statistical features of velocity and acceleration include mean $\bar{v}_i, \bar{a}_i$ and standard deviation $\sigma(v_i), \sigma(a_i)$.

Jerk, as the time derivative of acceleration, reflects driving smoothness:
\begin{equation}
j_i^t = \frac{a_i^{t+1} - a_i^{t-1}}{2\Delta t},
\end{equation}
\begin{equation}
\sigma(j_i) = \sqrt{\frac{1}{T_i} \sum_t (j_i^t - \bar{j}_i)^2}.
\end{equation}
A smaller $\sigma(j_i)$ indicates smooth acceleration changes and more comfortable driving.

\subsubsection{Temporal Feature Vector Construction}

For each trajectory $\tau_i$, we construct a temporal feature sequence:

\begin{equation}
\mathbf{F}_i = \{\mathbf{f}_i^t\}_{t=1}^{T_i},
\end{equation}
where the feature vector at each time step $\mathbf{f}_i^t$ contains:

\begin{itemize}
\item \textbf{Position Information}: World coordinates $(X_i^t, Y_i^t)$.
\item \textbf{Velocity Information}: Scalar velocity $v_i^t$ and components $(v_{x,i}^t, v_{y,i}^t)$.
\item \textbf{Acceleration Information}: Scalar acceleration $a_i^t$ and components $(a_{x,i}^t, a_{y,i}^t)$.
\item \textbf{Jerk Information}: Jerk $j_i^t$.
\item \textbf{Lane Information}: Current lane identifier $l_i^t$.
\item \textbf{Signal Information}: Inferred signal state for the current approach $\text{Signal}_i^t$.
\end{itemize}

Additionally, we compute statistical features for the entire trajectory:

\begin{equation}
f_i^{\text{stats}} = [\bar{v}_i, \sigma(v_i), \bar{a}_i, \sigma(a_i), \sigma(j_i), N_{\text{LC}}^i],
\end{equation}
where $N_{\text{LC}}^i$ is the number of lane changes. These features, along with trajectory temporal information, are converted to JSON format as input to the LLM.

\subsection{Rule Discovery Stage}

The rule discovery stage leverages the zero-shot reasoning capability of LLMs to induce behavioral differences from labeled AV and HDV trajectory data. 
We partition the labeled data into an AV set $\mathcal{D}_{\text{AV}}$ and a HDV set $\mathcal{D}_{\text{HDV}}$ to guide the model in extracting behavioral rules from normal driving behavior.
The model systematically compares behavioral differences between the two types of vehicles across the following dimensions:

\begin{itemize}
\item \textbf{Speed Control Patterns}: Speed fluctuation amplitude, change frequency, and adjustment strategies under different traffic densities.
\item \textbf{Acceleration Smoothness}: Jerk standard deviation, smoothness of acceleration and deceleration processes, and frequency of sudden braking.
\item \textbf{Lane-Change Behavior}: Triggering conditions, execution patterns, advance time, and speed adjustment strategies.
\item \textbf{Interaction Behavior}: Following distance maintenance strategies, and response characteristics to leading vehicle changes.
\item \textbf{Signal Response}: Deceleration patterns when approaching traffic signals, stopping, and starting behavior.
\item \textbf{Lane-Specific Behavior}: Behavioral differences across different lane types.
\end{itemize}

The model outputs a structured set of behavioral rules, divided into two categories:

\begin{itemize}
\item \textbf{Numerical Rules} $\mathcal{R}_{\text{numerical}}$: Decision rules based on quantified feature thresholds, such as jerk standard deviation below a threshold, average speed within a specific range, lateral acceleration during lane changes below a threshold, etc. Each numerical rule $r_k^{\text{num}} = \langle d_k, \phi_k^{\text{num}}, c_k \rangle$, where $\phi_k^{\text{num}}$ is a boolean decision function based on feature thresholds.
\item \textbf{Behavioral Statistical Rules} $\mathcal{R}_{\text{behavioral}}$: Decision rules based on behavioral pattern statistics, such as deceleration patterns before red lights (gradual vs. abrupt braking), lane-change decision timing (advanced planning vs. last-minute decisions), following distance consistency (stable maintenance vs. frequent adjustments), etc. Each behavioral statistical rule $r_k^{\text{beh}} = \langle d_k, \phi_k^{\text{beh}}, c_k \rangle$, where $\phi_k^{\text{beh}}$ makes judgments based on statistical characteristics of behavior sequences.
\end{itemize}

The rules output by the LLM undergo parsing and structuring, converting them into rule library entries. Through comparative analysis of the training set, the model generates an initial behavioral rule set $\mathcal{R} = \mathcal{R}_{\text{numerical}} \cup \mathcal{R}_{\text{behavioral}}$.

\subsection{Verification and Refinement Stage}

The verification and refinement stage tests the performance of rules on the vehicle identification task using an independent validation set, identifies failure cases, and refines rules through a reflection mechanism. 
For each rule $r_k$ in the rule library $\mathcal{R}$, we execute this prediction task on the validation set
$\mathcal{D}_{\text{val}}$ and compute $\text{Confidence}(r_k)$.
For prediction failure cases, we construct a failure case set $\mathcal{F}_k = \{(\tau_i, y_i, \hat{y}_i) \mid \phi_k(\tau_i) \neq y_i\}$ and feed back to the model for in-depth analysis.
The model reflects on the fundamental reasons for rule failures from the following perspectives:

\begin{itemize}
\item Whether the current threshold settings are reasonable and whether there are boundary conditions that need dynamic adjustment.
\item Whether the applicable scenarios of the rule are too broad or too restrictive and whether necessary contextual constraints are missing.
\item Whether a single feature is sufficient for discrimination or whether other features need to be combined to form composite rules.
\item Whether this rule has sufficient generalization capability and should be retained, modified, or deleted.
\end{itemize}

The model outputs specific refinement solutions, including recommended values for threshold adjustments, contextual constraints to be added, recommended feature combination methods, and complete expressions of refined rules.
The refinement operation is formalized as:
\begin{equation}
r_k' = \langle d_k', \phi_k', c_k' \rangle = \text{Refine}(r_k, \mathcal{F}_k, \text{LLM-Feedback}).
\end{equation}
The refined rules are retested, iterating until convergence. Finally, a verified high-confidence rule set $\mathcal{R}^*$ is obtained.

\subsection{Vehicle Type Identification Task}

The vehicle type identification task similarly employs two evaluation methods:

\begin{itemize}
    \item The first method uses the LLM to make comprehensive judgments based on numerical rules and behavioral statistical rules. For the trajectory $\tau_i$ to be identified, the model analyzes the matching of this trajectory across various rules:
\begin{equation}
\hat{y}_i^{\text{LLM}} = \text{LLM}(\tau_i, \mathcal{R}^*_{\text{numerical}}, \mathcal{R}^*_{\text{behavioral}}).
\end{equation}
The model comprehensively considers the quantitative judgment results of numerical rules and the pattern analysis results of behavioral statistical rules, providing final vehicle type judgments based on rule consistency, applicability scenario matching, and rule confidence.
    \item The second method individually evaluates the discriminative ability of each rule. For each rule $r_k \in \mathcal{R}^*$, its classification performance on the test set is independently judged:
\begin{equation}
\hat{y}_i^{(k)} = \begin{cases}
\text{AV}, & \text{if } \phi_k(\tau_i) = 1, \\
\text{HDV}, & \text{otherwise}.
\end{cases}
\end{equation}
By evaluating the accuracy, recall, and F1 scores of each rule, the most discriminative rules are identified.
\end{itemize}

\subsection{Algorithmic Workflow}

The overall algorithmic workflow of the SVBRD-LLM framework is summarized in 
Algorithm~\ref{alg:saba-llm}. The framework proceeds through four sequential stages. 
In Stage 1, raw video footage is processed by YOLOv26 and ByteTrack to extract vehicle 
trajectories, from which kinematic features (velocity, acceleration, jerk) and 
contextual attributes (lane identity, signal state) are computed and serialized into 
JSON format as LLM input. 
In Stage 2, the labeled training data are partitioned into AV and HDV subsets, and 
GPT-5 is prompted to perform comparative analysis across six behavioral dimensions, 
generating an initial rule set $\mathcal{R} = \mathcal{R}_{\text{numerical}} \cup 
\mathcal{R}_{\text{behavioral}}$ comprising 26 candidate rules. 
In Stage 3, each rule is evaluated on the validation set via lane-change prediction 
and AV identification tasks; rules below the confidence threshold $\theta$ trigger 
a reflection loop in which failure cases are fed back to the LLM for diagnosis and 
refinement, iterating until convergence to yield the final verified rule library 
$\mathcal{R}^*$. 
In Stage 4, the generalization capability of $\mathcal{R}^*$ is assessed on the 
held-out test set under two evaluation protocols: LLM-based comprehensive judgment 
and individual rule performance evaluation.

\begin{algorithm}
\caption{SVBRD-LLM Framework Workflow}
\label{alg:saba-llm}
\begin{algorithmic}[1]
\REQUIRE Traffic videos $\mathcal{V}$, labeled data $\mathcal{D}_{\text{labeled}}$, LLM
\ENSURE High-confidence rules $\mathcal{R}^*$, predictions $\{\hat{y}_i\}$

\STATE \textbf{Stage 1: Trajectory Extraction}
\FOR{each video $V \in \mathcal{V}$}
    \STATE Extract trajectories $\{\tau_i\}$ via YOLOv26 and ByteTrack
    \STATE Compute kinematic features $\{f_i\}$
\ENDFOR
\STATE Split data: $\mathcal{D}_{\text{train}}, \mathcal{D}_{\text{val}}, \mathcal{D}_{\text{test}}$

\STATE \textbf{Stage 2: Rule Discovery}
\STATE Separate $\mathcal{D}_{\text{train}}$ into $\mathcal{D}_{\text{AV}}$ and $\mathcal{D}_{\text{HDV}}$
\STATE LLM performs comparative analysis of AV and HDV driving behavior
\STATE Generate and parse rule set $\mathcal{R} = \mathcal{R}_{\text{numerical}} \cup \mathcal{R}_{\text{behavioral}}$

\STATE \textbf{Stage 3: Verification and Refinement}
\REPEAT
    \FOR{each rule $r_k \in \mathcal{R}$}
        \STATE Test performance on vehicle identification task on $\mathcal{D}_{\text{val}}$
        \STATE Compute Confidence$(r_k)$
        \IF{Confidence$(r_k) < \theta$}
            \STATE Collect failure cases $\mathcal{F}_k$
            \STATE $r_k \leftarrow$ Refine via LLM reflection
        \ENDIF
    \ENDFOR
\UNTIL{convergence}
\STATE $\mathcal{R}^* \leftarrow \{r_k \mid \text{Confidence}(r_k) \geq \theta\}$

\STATE \textbf{Stage 4: Application Evaluation}
\FOR{each $\tau_i \in \mathcal{D}_{\text{test}}$}
    \STATE \textbf{Task (Vehicle Identification):}
    \STATE \quad Method 1: LLM comprehensive rule-based judgment $\hat{y}_i^{\text{LLM}}$
    \STATE \quad Method 2: Individual rule performance evaluation $\hat{y}_i^{(k)}$
\ENDFOR
\RETURN $\mathcal{R}^*$, prediction results
\end{algorithmic}
\end{algorithm}

\section{Experiments}
\label{sec:3}

\subsection{Experimental Setup}

\begin{figure}[t!]
\centerline{\includegraphics[width=\columnwidth]{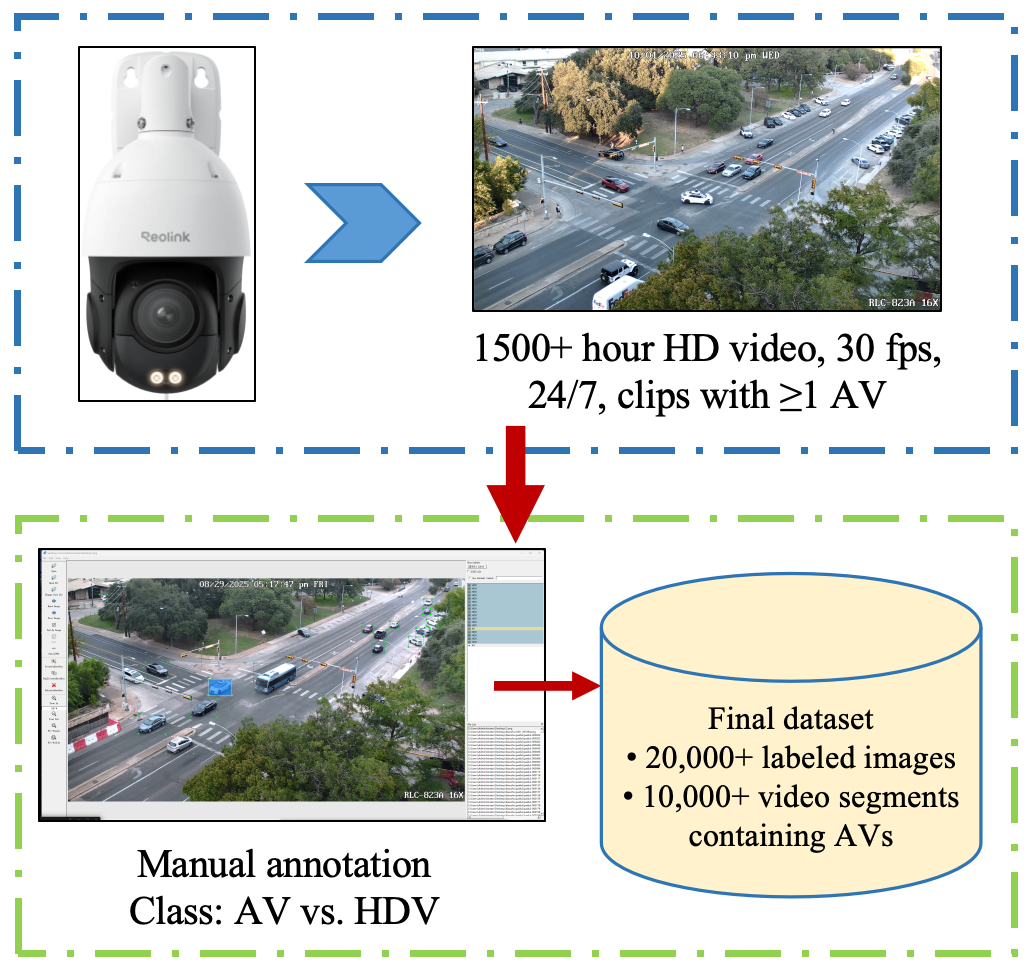}}
\caption{Roadside video dataset collection.}
\label{fig:dataset}
\end{figure}

We validate our framework using a new roadside video dataset in the commercial operating area of Waymo, as illustrated in Figure~\ref{fig:dataset}. 
A Reolink RLC-823A 16X PTZ camera is used to overlook a public signalized multi-lane intersection. 
The intersection features multiple approach lanes with left-turn, through, and right-turn movements, capturing diverse vehicle interactions including car-following, lane-changing, and turning maneuvers under varying traffic signal phases. 
The camera records HD video at 30 fps, operating continuously in both daytime color and nighttime infrared modes. 
We collect more than 1,500 hours of footage. 
From this continuous recording, we extract all clips containing at least one visually identifiable Waymo vehicle, characterized by its distinctive rooftop LiDAR sensor dome. 
This filtering process yields approximately 15 hours of relevant footage for subsequent analysis.
Each video segment represents a continuous event of a vehicle passing through the intersection. 
These segments are processed through YOLOv26 detection and ByteTrack tracking to extract complete trajectories for behavioral analysis. 
Vehicle labels are obtained through a semi-automatic pipeline: YOLOv26 detects all vehicles in each frame, ByteTrack associates detections across frames to form trajectories, and Waymo vehicles are automatically flagged based on visual features (sensor dome and vehicle livery) and subsequently verifies through manual review to ensure labeling accuracy. 
The complete dataset sizes and splits for each task are summarized in Table~\ref{tab:dataset_splits}.

\begin{table*}[t]
\caption{Dataset splits for the AV identification task.}
\label{tab:dataset_splits}
\centering
\begin{tabularx}{\textwidth}{@{}l l >{\centering\arraybackslash}X >{\centering\arraybackslash}X >{\centering\arraybackslash}X >{\centering\arraybackslash}X@{}}
\toprule
\textbf{Task} & \textbf{Data Type} & \textbf{Total} & \textbf{Training} & \textbf{Validation} & \textbf{Test} \\
\midrule
AV Identification & Video Segments & 10,913 & 6,548 & 2,182 & 2,183 \\

\bottomrule
\end{tabularx}
\end{table*}

\subsubsection{Implementation Details} 

\begin{figure}[t]
\centerline{\includegraphics[width=\columnwidth]{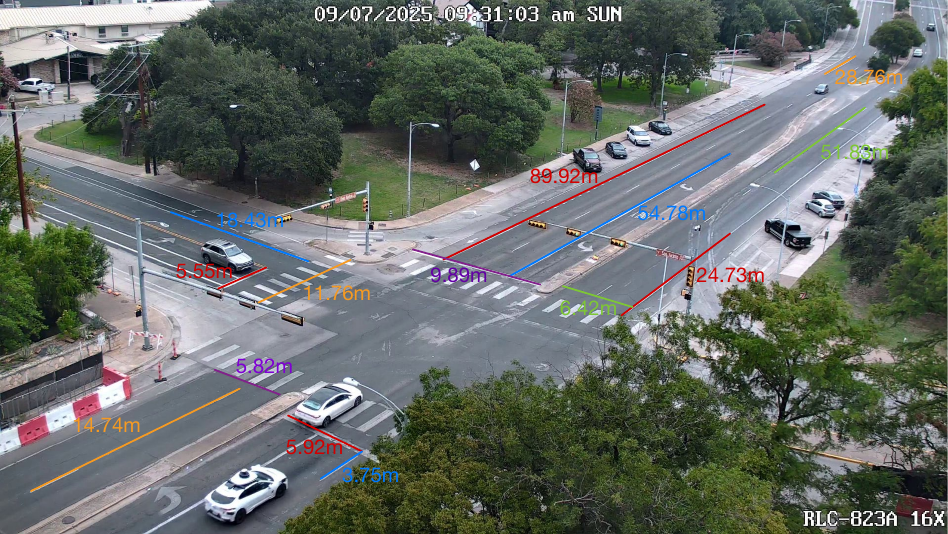}}
\caption{Spatial calibration using known reference dimensions to establish pixel-to-meter ratio for computing physical kinematic features.}
\label{fig:calibration}
\end{figure}

We employ YOLOv26 for vehicle detection with a confidence threshold of 0.5, and ByteTrack for multi-object tracking with IOU threshold 0.5 and maximum disappearance frames set to 30. 
To convert pixel coordinates to physical units, we perform spatial calibration using known reference dimensions in the scene, as illustrated in Figure~\ref{fig:calibration}. 
This calibration establishes a pixel-to-meter ratio, enabling the computation of real-world velocity (m/s), acceleration (m/s²), and jerk (m/s³) from trajectory data.

In the rule discovery and verification stages, we concatenate selected trajectory information from the training set with prompts and input them into GPT-5 for deep analysis, thereby generating behavioral rules. 
Model parameters: temperature 0.7, maximum output tokens 2,000.

\subsubsection{Evaluation Metrics} We use Accuracy, Precision, Recall, F1-score, and ROC-AUC to evaluate model performance.

\begin{table*}[t]
\caption{Complete verified behavioral rule library with individual accuracy on the test set.}
\label{tab:rules}
\centering
\footnotesize
\setlength{\tabcolsep}{4pt}
\renewcommand{\arraystretch}{0.88}
\begin{tabularx}{\textwidth}{@{}c >{\raggedright\arraybackslash}X >{\raggedright\arraybackslash}p{2.6cm} c c c@{}}
\toprule
\textbf{ID} & \textbf{Description} & \textbf{Threshold / Pattern}
  & \textbf{Category} & \textbf{AV vs.\ HDV} & \textbf{Indiv.\ Acc.} \\
\midrule
\multicolumn{6}{c}{\textit{Numerical Rules (N\,=\,13)}} \\
\midrule

R1  & AV acceleration variation smaller
    & $\sigma(a) < 0.35$~m/s$^{2}$
    & Speed
    & $0.31$ vs.\ $0.53$~m/s$^{2}$
    & 70\% \\

R2  & AV deceleration variation smaller
    & $\sigma(\text{decel}) < 0.45$~m/s$^{2}$
    & Speed
    & $0.38$ vs.\ $0.66$~m/s$^{2}$
    & 66\% \\

R3  & AV maintains constant low speed near queue
    & 5--15~km/h (queue approach)
    & Speed
    & Constant vs.\ Variable
    & 62\% \\

R4  & AV maintains larger time headway
    & THW $> 2.5$~s
    & Car-following
    & $3.1$ vs.\ $1.8$~s
    & 70\% \\

R5  & AV decelerates proactively before lane change
    & $0.40$--$0.70$~m/s$^{2}$
    & Lane change
    & $0.52$ vs.\ $0.14$~m/s$^{2}$
    & 74\% \\

R6  & AV executes lane change at shallower angle
    & $10°$--$18°$
    & Lane change
    & $14°$ vs.\ $24°$
    & 63\% \\

R7  & AV lane change duration longer
    & $3.0$--$5.5$~s
    & Lane change
    & $3.8$ vs.\ $2.4$~s
    & 68\% \\

R8  & AV lateral velocity during lane change smaller
    & $|v_y| < 0.60$~m/s
    & Lane change
    & $0.42$ vs.\ $0.71$~m/s
    & 67\% \\

R9  & AV speed variation during lane change smaller
    & $\Delta v < 2.5$~m/s
    & Lane change
    & $1.8$ vs.\ $3.3$~m/s
    & 71\% \\

R10 & AV acceleration variation stable in car-following
    & $\Delta a < 0.50$~m/s$^{2}$
    & Car-following
    & $0.38$ vs.\ $0.65$~m/s$^{2}$
    & 73\% \\

R11 & AV lower jerk std.\ deviation (overall)
    & $\sigma(\text{jerk}) < 0.80$~m/s$^{3}$
    & Smoothness
    & $0.65$ vs.\ $1.25$~m/s$^{3}$
    & 77\% \\

R12 & AV lower jerk std.\ deviation (acceleration phase)
    & $\sigma(\text{jerk}) < 0.70$~m/s$^{3}$
    & Smoothness
    & $0.58$ vs.\ $1.09$~m/s$^{3}$
    & 74\% \\

R13 & AV lower jerk std.\ deviation (deceleration phase)
    & $\sigma(\text{jerk}) < 1.20$~m/s$^{3}$
    & Smoothness
    & $0.92$ vs.\ $1.58$~m/s$^{3}$
    & 75\% \\

\midrule
\multicolumn{6}{c}{\textit{Behavioral Statistical Rules (N\,=\,7)}} \\
\midrule

R14 & AV rarely performs hard braking
    & $\leq\!1$ event/trip (decel $> 3.0$~m/s$^{2}$)
    & Smoothness
    & $0.3$ vs.\ $2.1$ events/trip
    & 76\% \\

R15 & AV rarely performs hard acceleration
    & $\leq\!1$ event/trip (accel $> 2.5$~m/s$^{2}$)
    & Smoothness
    & $0.4$ vs.\ $1.8$ events/trip
    & 75\% \\

R16 & AV maintains stable speed throughout trajectory
    & Low speed-fluctuation pattern
    & Speed
    & Stable vs.\ Fluctuating
    & 68\% \\

R17 & AV exhibits conservative behavior at signals
    & No yellow- or red-light rushing
    & Signal response
    & Conservative vs.\ Aggressive
    & 79\% \\

R18 & AV starts slower but more consistently
    & Gradual, uniform acceleration pattern
    & Starting
    & Slow-consistent vs.\ Fast-varied
    & 64\% \\

R19 & AV maintains stable lateral position
    & Minimal lateral-position wandering
    & Lane keeping
    & Stable vs.\ Variable
    & 61\% \\

R20 & AV maintains consistent car-following distance
    & Low THW-variation pattern
    & Car-following
    & Consistent vs.\ Variable
    & 70\% \\

\bottomrule
\end{tabularx}
\end{table*}

We compare the proposed \textbf{SVBRD-LLM} against both traditional interpretable models and deep learning baselines. 
All baseline models use the same kinematic features (i.e., mean velocity, velocity standard deviation, mean acceleration, acceleration standard deviation, jerk standard deviation, and lane-change count) as input.

\begin{itemize}
    \item \textbf{Logistic Regression (LR)}: A linear classifier trained with L2 
    regularization (C=1.0) using the scikit-learn implementation \cite{pedregosa2011scikit}. 
    This serves as a simple interpretable baseline.

    \item \textbf{Decision Tree (DT)}: A tree-based classifier with maximum depth of 10 
    and minimum samples per leaf of 5, providing rule-based interpretability. 
    We use the CART algorithm \cite{breiman1984classification} implemented in 
    scikit-learn \cite{pedregosa2011scikit}.

    \item \textbf{Random Forest (RF)}: An ensemble of 100 decision trees 
    \cite{breiman2001random} with maximum depth of 15, combining multiple weak learners 
    for improved performance while maintaining partial interpretability through 
    feature importance analysis.

    \item \textbf{Long Short-Term Memory (LSTM)}: A recurrent neural network baseline 
    \cite{hochreiter1997long} consisting of 2 stacked LSTM layers with 128 hidden units 
    each, followed by a fully connected output layer with softmax activation. The model takes raw trajectory sequences (i.e., position, velocity, acceleration) 
    as input. Training is performed using Adam optimizer \cite{kingma2014adam} with a 
    learning rate of 0.001 for 100 epochs, with early stopping based on validation loss.

    \item \textbf{SVBRD-LLM (Ours)}: The proposed framework using GPT-5 zero-shot 
    prompting for rule discovery, with temperature 0.7 and maximum output tokens 2,000. 
    Unlike other baselines, our \textbf{SVBRD-LLM} generates explicit behavioral rules 
    with semantic descriptions and contextual applicability conditions.
\end{itemize}

\subsubsection{Validation of Traffic Signal State Inference}

Since direct ground-truth signal annotations are unavailable from the 
camera view, we validate the inference indirectly: absent red-light 
violations, a vehicle stopping near the stop line provides behavioral 
confirmation of a red phase, and a lead vehicle clearing the stop line 
confirms green. Under this assumption, the inference accuracy approaches 
100\% in practice.

\begin{figure*}[t]
\centerline{\includegraphics[width=1\linewidth]{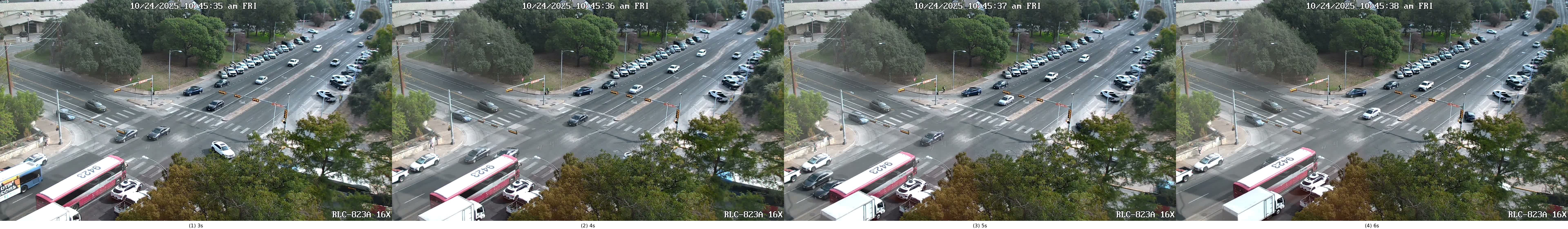}}
\caption{Illustration of traffic signal state inference validation for 
the WB through lanes. Frames at $t = 3$\,s and $t = 4$\,s show vehicles 
stationary near the crosswalk under a red signal; at $t = 5$\,s the 
signal turns green and vehicles begin to move.}
\label{fig:signal_gt}
\end{figure*}

Figure~\ref{fig:signal_gt} illustrates a representative example for the 
Westbound~(WB) through lanes. At $t = 3$\,s and $t = 4$\,s, the signal 
is red and most vehicles remain stationary near the crosswalk. At 
$t = 5$\,s, the signal turns green and vehicles begin to move, with 
velocities increasing clearly in the extracted trajectories. This 
behavioral consistency confirms the correctness of the inferred signal 
states.

\subsection{Prompt Process}

This Prompt Process demonstrates the end-to-end operation of SVBRD-LLM
across three stages using trajectory samples collected at the monitored
signalized intersection in Austin, TX\@.
Four samples are selected: two Waymo AV samples and
two HDV samples, all traveling along the
eastbound through lanes under red-light approach and green-light
car-following conditions, respectively.

In Stage~1, the four samples are organized according to the prompt template illustrated in Figure~\ref{fig:case_prompt}.
Trajectory data are formatted as JSON objects and submitted to GPT-5
together with the role description, scene context, and two-step
reasoning instructions.
The model first performs a step-by-step comparative analysis of AV
and HDV behavior across five dimensions, then outputs structured
rule hypotheses in JSON format.
No camera imagery is provided at any stage; all reasoning is performed
exclusively on structured trajectory and kinematic feature data.

\subsection{Rule Discovery and Verification}

GPT-5 analyzes trajectory data from the training set and generates 26 initial behavioral rules, comprising 16 numerical rules and 10 behavioral statistical rules covering speed control, lane-change decision-making, car-following behavior, and acceleration smoothness.
Table~\ref{tab:rules} presents the complete verified rule library, covering all 20 high-confidence rules across both categories.
These rules quantify essential behavioral differences between AVs and HDVs through quantitative thresholds and behavioral pattern statistics.

The numerical rules capture quantifiable differences such as AVs exhibiting more linear acceleration with lower standard deviation  (R1, $\sigma$(a): 0.31 vs.\ 0.53~m/s²), smoother deceleration variability (R2, $\sigma$(\text{decel}): 0.38 vs.\ 0.64~m/s²), maintaining consistent low-speed creeping near intersections (R3, 5--15~km/h), proactive deceleration before lane changes (R5: 0.52 vs.\ 0.14~m/s²), smoother lane-change execution angles (R6: 14° vs.\ 24°), smaller speed variation during lane changes (R9: 1.8 vs.\ 3.3~m/s), and significantly lower jerk standard deviation across all driving phases (R11--R13, $\sigma$(\text{jerk}): 0.65 vs.\ 1.22~m/s³), reflecting the
motion-smoothness advantage of autonomous driving systems. In addition, AVs maintain significantly larger time headways during car-following (R4: $3.1$ vs.\ $1.8$~s), execute lane changes over longer durations with smaller lateral velocities (R7: $3.8$ vs.\ $2.4$~s; R8: $|v_y|$ $0.42$ vs.\ $0.71$~m/s), 
and exhibit lower jerk variability specifically during the acceleration phase (R12: $0.58$ vs.\ $1.09$~m/s$^{3}$), collectively reinforcing the pattern of conservative and smooth motion planning characteristic of autonomous systems.

\begin{table*}[t]
\caption{Refinement examples for low-accuracy rules.}
\label{tab:refinement}
\centering
\begin{tabular*}{\textwidth}{@{\extracolsep{\fill}}cccc@{}}
\toprule
\textbf{ID} & \textbf{Before Refinement} & \textbf{After Refinement} & \textbf{Reason} \\
\midrule
R3  & Constant speed 0--10~km/h
    & 5--15~km/h (queue approach, non-congested only)
    & Exclude stationary state; cover low-speed creep range \\
R9  & $\Delta v < 1.0$~m/s
    & $\Delta v < 2.5$~m/s (non-congested only)
    & Allow reasonable speed variation during lane change \\
R13 & $\sigma(\text{jerk}) < 0.80$~m/s$^3$
    & $\sigma(\text{jerk}) < 1.20$~m/s$^3$ (deceleration phase only)
    & Accommodate higher-speed deceleration scenarios \\
\bottomrule
\end{tabular*}
\end{table*}

\begin{figure}[t]
\centerline{\includegraphics[width=\columnwidth]{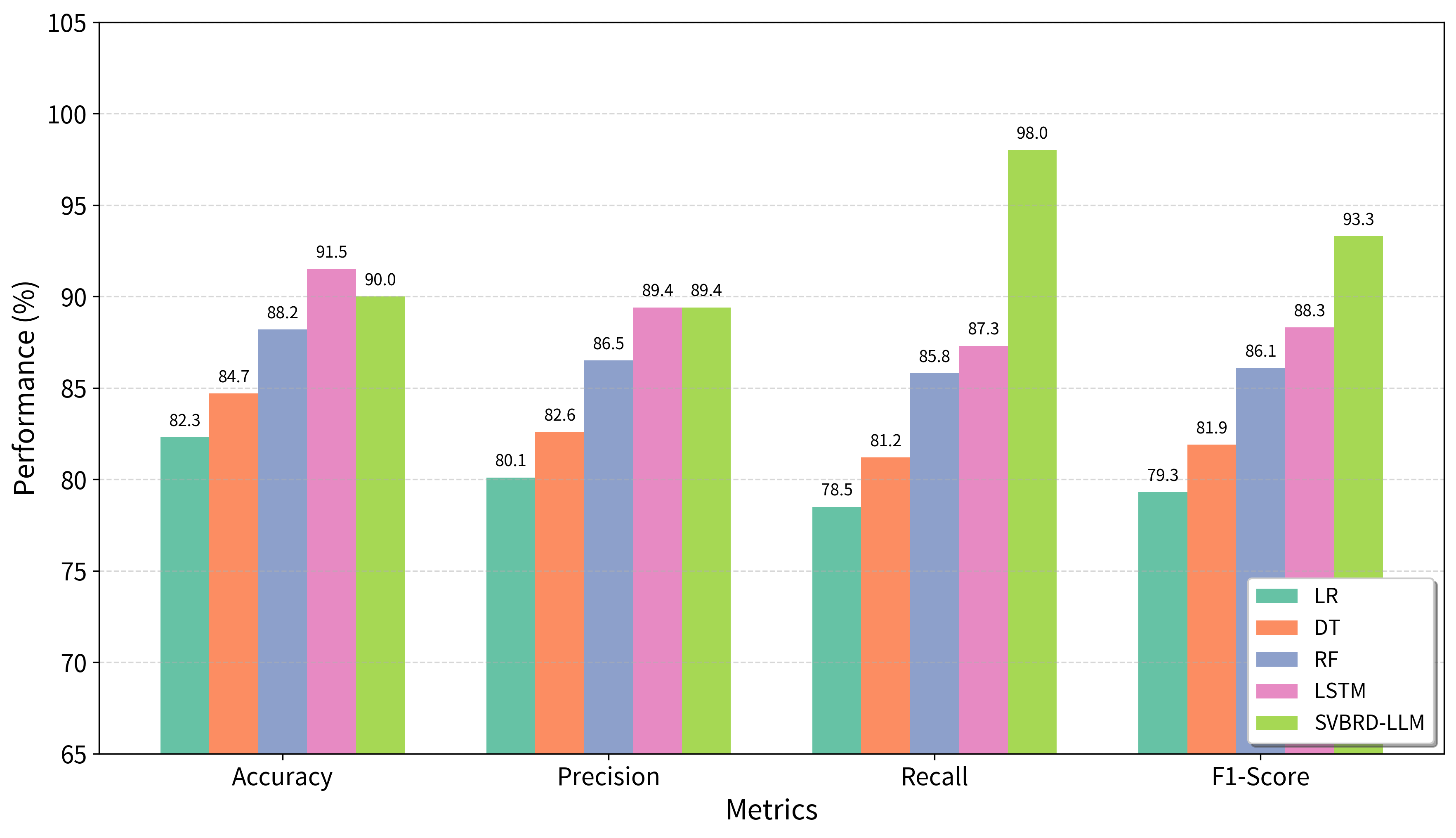}}
\caption{Performance comparison of different methods on AV identification task.}
\label{fig:task_comparison}
\end{figure}

The behavioral statistical rules identify pattern-based differences: AVs maintain more stable speeds with lower fluctuation (R16), exhibit conservative behavior at traffic signals without attempting to rush through yellow or red lights (R17), demonstrate slower but more consistent starting behavior (R18), and maintain more stable lateral positions during lane-keeping (R19). Furthermore, AVs generate significantly fewer hard-braking and hard-acceleration events per trip (R14: $0.3$ vs.\ $2.1$ events; R15: $0.4$ vs.\ $1.8$ events), and sustain more consistent car-following distances with lower time-headway variation (R20), reflecting the systematic smoothness and conservatism of algorithmic vehicle control.

In the verification stage, these 35 rules are tested on the validation set. 
Initial testing reveals that 11 rules achieve individual accuracy above 70\%, 16 rules between 60--70\%, and 3 rules below 60\%. 
Failure case analysis identifies three key issues: neglecting traffic density context, overly strict thresholds, and insufficient scenario applicability.
Table~\ref{tab:refinement} presents typical refinement examples, including adding contextual constraints (e.g., ``applicable only in non-congested scenarios'') and relaxing threshold ranges (e.g., expanding lane change speed variation from $\Delta v < 1.0$~m/s to $\Delta v < 2.5$~m/s, and adjusting jerk threshold from $\sigma(\text{jerk}) < 0.80$~m/s³ to $\sigma(\text{jerk}) < 1.20$~m/s³) to improve rule generalization. 
After two iterations, 20 high-confidence rules are retained with an average individual validation accuracy of 70.3\%.

\subsection{AV Identification Task Evaluation}

We evaluate the predictive power of discovered rules on the test set through the AV identification downstream task. 
Figure~\ref{fig:task_comparison} illustrates the performance comparison across all methods.

Traditional methods (i.e., LR, DT, RF, and LSTM) take as input 5-second historical trajectories of the target vehicle and its neighboring vehicles, including kinematic feature sequences such as position, velocity, and acceleration. 
Our \textbf{SVBRD-LLM} takes three types of input: 

\begin{itemize}
    \item Statistical features extracted from trajectories. 
    \item Contextual information (e.g., lane position, traffic signal state, surrounding vehicle distribution, etc.). 
    \item Verified numerical rules and behavioral statistical rules. 
\end{itemize}

The LLM performs comprehensive reasoning based on this information.



\textit{Autonomous Vehicle Identification} task identifies whether an observed 
vehicle is autonomous or human-driven. 
As illustrated in Figure~\ref{fig:task_comparison}, LSTM achieves the highest 
accuracy (91.5\%), demonstrating the advantage of end-to-end learning in 
capturing complex behavioral patterns. 
\textbf{SVBRD-LLM} achieves 90.0\% accuracy while significantly leading all 
methods in F1-score (93.3\%) and recall (98.0\%). 
This performance characteristic stems from the comprehensive capture of AV 
behavioral features during the rule discovery process: rules extracted through 
comparative analysis of AV and HDV trajectory data can effectively identify 
typical AV behavioral patterns, achieving an extremely low miss rate (only 2\% 
of AVs are missed). 
Such high recall is particularly valuable in safety-critical applications---such 
as regulatory monitoring and incident investigation---where failing to detect an 
AV carries greater operational risk than an occasional false positive. 
Meanwhile, some HDVs exhibiting similar features are misclassified as AVs, 
resulting in precision comparable to LSTM. 
Among traditional methods, RF performs best with 88.2\% accuracy.

Beyond quantitative performance, the rules generated by \textbf{SVBRD-LLM} 
offer a key advantage that purely data-driven methods cannot provide: 
\emph{interpretability and auditability}. 
Each rule carries explicit contextual constraints and semantic descriptions---for 
example, ``under non-congested traffic conditions, AVs exhibit smoother 
deceleration with jerk standard deviation below 0.80~m/s$^3$''---enabling 
post-hoc verification of individual predictions and transparent audit trails for 
regulatory compliance. 
The broader implications of these discovered rules for traffic safety and 
regulatory policy are further discussed in Section~\ref{sec:insights}.

\subsection{Ablation Study}

\begin{figure}[t]
\centerline{\includegraphics[width=\linewidth]{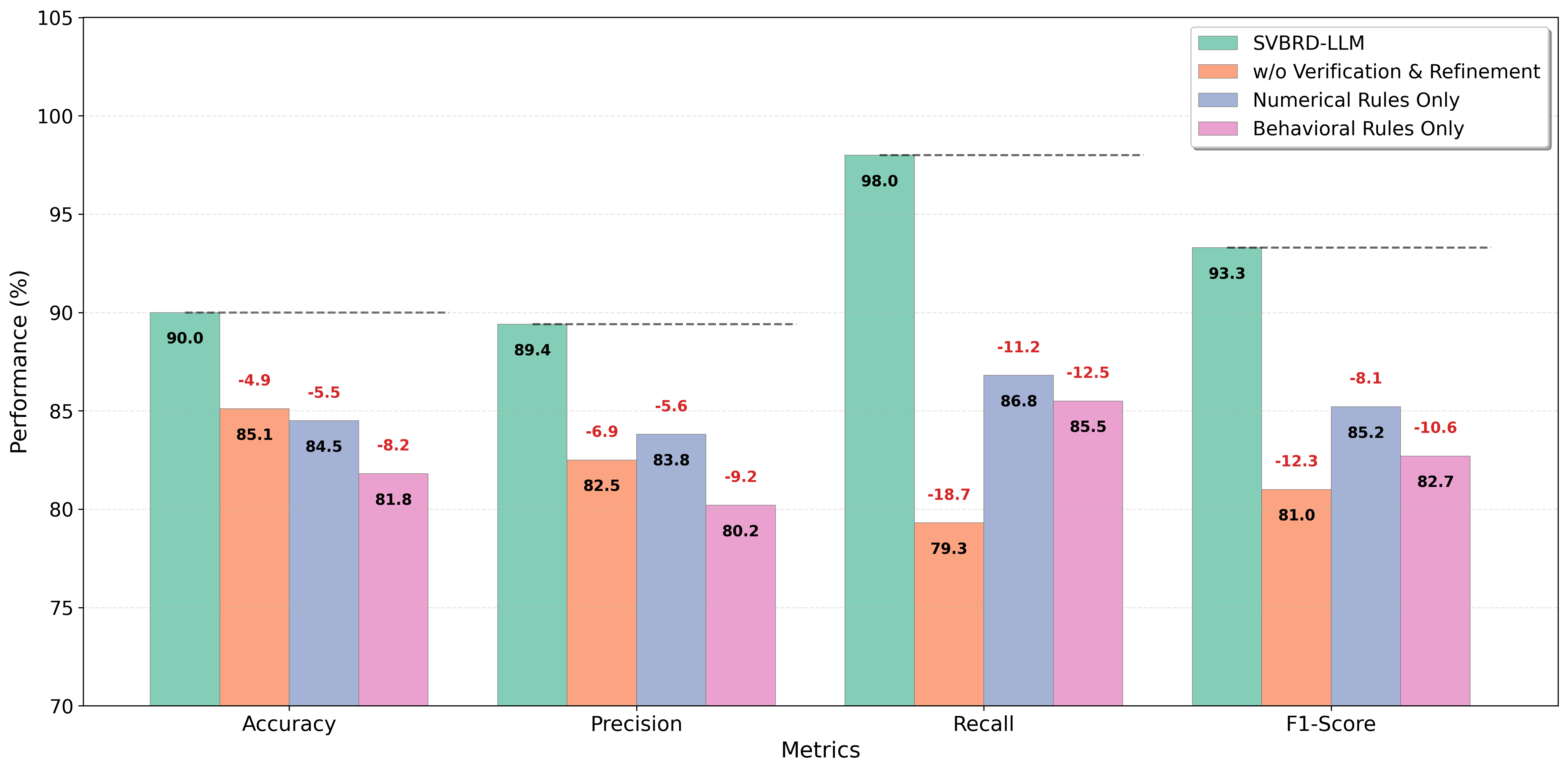}}
\caption{Ablation study results on AV identification task. The dashed lines indicate the baseline performance of the complete SVBRD-LLM method. Numbers above bars show the performance decrease (-X.X) relative to the complete method, while numbers within bars show the actual performance values.}
\label{fig:ablation}
\end{figure}

The ablation study validates the necessity of each component in the framework on the AV identification task. 
Figure~\ref{fig:ablation} illustrates the performance comparison across different configurations.

As can be seen in the figure, the complete method achieves the best performance across all metrics, while removing any component leads to performance degradation. 
Removing the verification and refinement mechanism causes the most significant performance drop, with accuracy decreasing by 4.9\% points and F1-score decreasing by 12.3\% points. 
This demonstrates that iterative reflection and rule refinement are crucial for filtering spurious correlations and improving rule reliability. 
The mechanism enhances rule generalization capability by testing rules on the validation set, identifying failure cases, and performing targeted improvements.

Comparing rule categories, the 13 numerical rules alone achieve 84.5\% accuracy and 85.2\% F1-score, while the 7 behavioral statistical rules alone achieve 81.8\% accuracy and 82.7\% F1-score.
This indicates that both rule types provide complementary discriminative information: numerical rules provide precise boundary judgments through quantitative thresholds (such as jerk standard deviation, deceleration magnitude), while behavioral statistical rules capture qualitative driving style characteristics through pattern recognition (such as speed stability, conservative signal response). 
The combination of both rule types enables the framework to comprehensively characterize vehicle behavioral features from multiple dimensions.

\subsection{Insights for Traffic Safety and Regulation}
\label{sec:insights}

Based on the 20 verified high-confidence rules, Table~\ref{tab:insights} summarizes the key behavioral differences between AVs and HDVs discovered through our framework, integrating both numerical thresholds and behavioral patterns.
These quantitative findings and behavioral patterns provide important insights for traffic management, safety assessment, and regulatory decision-making.

\begin{table*}[t]
\caption{Key behavioral differences between autonomous vehicles and human-driven vehicles}
\label{tab:insights}
\centering
\begin{tabular*}{\textwidth}{@{\extracolsep{\fill}}lccc@{}}
\toprule
\textbf{Behavioral Metrics} & \textbf{Autonomous Vehicle (AV)} & \textbf{Human-Driven Vehicle (HDV)} & \textbf{Difference} \\
\midrule
\multicolumn{4}{c}{\textit{Numerical Characteristics}} \\
\midrule
Acceleration Std.\ Dev.\ (m/s²)   & 0.31 & 0.53 & $-42\%$ \\
Deceleration Std.\ Dev.\ (m/s²)   & 0.38 & 0.64 & $-41\%$ \\
Jerk Std.\ Dev.\ (m/s³)           & 0.65 & 1.22 & $-47\%$ \\
Pre-LC Deceleration (m/s²)        & 0.52 & 0.14 & $+271\%$ \\
LC Execution Angle (°)            & 14   & 24   & $-42\%$ \\
\midrule
\multicolumn{4}{c}{\textit{Behavioral Patterns}} \\
\midrule
Speed Stability   & Stable          & Fluctuating & Consistent  \\
Signal Response   & Conservative    & Aggressive  & No rushing  \\
Starting Behavior & Slow-consistent & Fast-varied & Gradual     \\
Lane Keeping      & Stable lateral  & Variable    & Minimal wander \\
\bottomrule
\end{tabular*}
\end{table*}

\begin{itemize}
    \item \textbf{Motion Smoothness and Safety}: AVs demonstrate 42\% lower acceleration standard deviation (0.31 vs.\ 0.53~m/s²) and 47\% lower jerk standard deviation (0.65 vs.\ 1.22~m/s³), combined with stable speed maintenance patterns, directly reducing rear-end collision risks through more predictable motion. 
    The substantially reduced jerk also improves passenger comfort and lowers energy consumption. 
    The consistent speed stability pattern, contrasting with HDVs' fluctuating behavior, enables smoother mixed-traffic flow.
    \item \textbf{Conservative Behavior and Traffic Flow}: AVs exhibit conservative patterns across multiple dimensions: approximately 3$\times$ higher pre-deceleration during lane changes (0.52 vs.\ 0.14~m/s², $+$271\%) with 42\% smaller execution angles (14° vs.\ 24°), no yellow or red light rushing at signalized intersections, and slower but more consistent starting patterns. 
    While these characteristics enhance safety, they also require larger gaps in dense traffic and longer startup times, informing the design of merging zones, on-ramps, signal timing strategies, and intersection management that explicitly account for AV behavioral characteristics.
    \item \textbf{Lane Discipline and Lateral Control}: AVs maintain stable lateral positions with minimal lane wandering during lane-keeping, contrasting with HDVs' variable lateral behavior. 
    This discipline improves lane capacity utilization and reduces side-swipe collision risks, but may require adjusted lane width standards and marking visibility requirements optimized for autonomous systems.
    \item \textbf{Regulatory Monitoring and Transparency}: The discovered rules, encompassing both numerical thresholds and behavioral patterns, provide a comprehensive foundation for AV performance benchmarks and compliance criteria. The high recall rate (98.0\%) ensures comprehensive monitoring for incident investigation and liability frameworks. 
    The rule-based approach offers transparent, auditable decisions with explicit contextual conditions (e.g., ``non-congested scenarios''), meeting regulatory explainability requirements and enabling post-hoc review of system decisions.
    \item \textbf{Adaptive Traffic Management}: The systematic behavioral differences between AVs and HDVs---spanning motion control, decision-making patterns, and interaction styles---necessitate adaptive strategies as penetration rates increase. 
    Traffic operations should develop monitoring systems to identify AV presence based on discovered behavioral signatures and adjust control measures (e.g., signal timing, speed limits, lane allocation) to optimize mixed traffic performance during the transition to higher automation levels.
\end{itemize}

\begin{figure*}[t]          
\centering
\begin{tcolorbox}[
  enhanced,                 
  width        = \textwidth,
  colback      = mainbg,
  colframe     = hdrtcol,
  title        = {\bfseries\small
                  SVBRD-LLM Behavioral Rule Discovery\,---\,Prompt Template},
  coltitle     = white,
  colbacktitle = hdrtcol,
  boxrule = 0.8pt, arc = 3pt,
  left=5pt, right=5pt, top=3pt, bottom=3pt
]

\noindent{\small\textbf{[Role Description]}}\\[1pt]
{\small You are an expert in traffic behavior analysis, specializing in
identifying interpretable behavioral differences between autonomous
vehicles~(AV) and human-driven vehicles~(HDV) from structured
trajectory data.\quad\textbf{No image input is provided at any stage.}}

\smallskip

\noindent{\small\textbf{[Scene Context]}}\\[1pt]
{\small Multi-lane signalized intersection, Austin\,TX
(roadside camera, 30\,fps, RLC-823A\,16X).
Lane structure: EB-Left, EB-Through-1\,(EB-T1), EB-Through-2\,(EB-T2),
EB-Right, and corresponding WB\,/\,NB\,/\,SB approaches.
Trajectory window: \textbf{5\,s}, sampled every \textbf{0.5\,s}
(11 time steps per sample).
Fields per step:
\texttt{t}\,(s),\;
\texttt{x,\,y}\,(m),\;
\texttt{vx,\,vy}\,(m/s),\;
\texttt{ax,\,ay}\,(m/s$^{2}$),\;
\texttt{jerk}\,(m/s$^{3}$),\;
\texttt{lane},\;
\texttt{signal},\;
\texttt{THW}\,(s).}

\smallskip

\begin{tcolorbox}[
  enhanced, width=\linewidth,
  colback=avboxbg, colframe=green!50!black,
  boxrule=0.5pt, arc=2pt,
  left=3pt, right=3pt, top=1pt, bottom=1pt
]
\tcbinputlisting{
  listing file    = {json_av.txt},
  listing only,
  listing options = {basicstyle=\ttfamily\scriptsize,
                     breaklines=true, keepspaces=true, columns=flexible}
}
\end{tcolorbox}

\vspace{2pt}

\begin{tcolorbox}[
  enhanced, width=\linewidth,
  colback=hdvboxbg, colframe=orange!55!black,
  boxrule=0.5pt, arc=2pt,
  left=3pt, right=3pt, top=1pt, bottom=1pt
]
\tcbinputlisting{
  listing file    = {json_hdv.txt},
  listing only,
  listing options = {basicstyle=\ttfamily\scriptsize,
                     breaklines=true, keepspaces=true, columns=flexible}
}
\end{tcolorbox}

\smallskip

\noindent{\small\textbf{[Reasoning Instructions]}}\\[1pt]
{\small\textbf{Step~1\,---\,Comparative Analysis.}
Compare AV vs.\ HDV behavior across the following five dimensions,
citing specific numerical evidence from the samples above:}
\begin{enumerate}[leftmargin=16pt, itemsep=0pt,
                  topsep=1pt, parsep=0pt, label=(\arabic*)]
  {\small
  \item Speed stability: $\bar{v}$, $\sigma(v)$, overall speed range.
  \item Jerk smoothness: $\sigma(\text{jerk})$,
        hard-braking / hard-acceleration count.
  \item Car-following THW: mean and $\sigma(\text{THW})$.
  \item Signal response: deceleration onset distance,
        yellow-/red-light rushing.
  \item Lateral stability: $\sigma_y$, lane-wandering pattern.}
\end{enumerate}

\vspace{2pt}
{\small\textbf{Step~2\,---\,Structured Rule Output.}
For each identified behavioral difference, output one rule entry
in the following JSON format, as illustrated by the examples below:}

\vspace{2pt}

\begin{tcolorbox}[
  enhanced, width=\linewidth,
  colback=instbg, colframe=gray!55,
  boxrule=0.5pt, arc=2pt,
  left=3pt, right=3pt, top=1pt, bottom=1pt
]
\tcbinputlisting{
  listing file    = {json_rules.txt},
  listing only,
  listing options = {basicstyle=\ttfamily\scriptsize,
                     breaklines=true, keepspaces=true, columns=flexible}
}
\end{tcolorbox}

\end{tcolorbox}
\caption{Structured prompt template used in stage2  and 3 of SVBRD-LLM\@.AV and HDV trajectory samples are provided in JSON format with 0.5\,s
sampling over a 5\,s window.}
\label{fig:case_prompt}
\end{figure*}




\section{Conclusion}
\label{sec:4}

This paper proposes the \textbf{SVBRD-LLM} framework, which automatically discovers, verifies, and applies interpretable behavioral rules from real traffic videos for AV identification through zero-shot prompt engineering. 
Experiments on over 1,500 hours of real-world traffic videos from Waymo's commercial operating area demonstrate that the framework achieves 90.0\% accuracy and 93.3\% F1-score in AV identification, with 98.0\% recall ensuring comprehensive monitoring. 
The discovered 20 high-confidence rules, comprising 13 numerical rules and 7 behavioral statistical rules, reveal distinctive characteristics of AVs in motion smoothness, conservative behavior, and lane discipline. 
Each rule is accompanied by semantic descriptions, quantitative thresholds or behavioral patterns, applicable contexts, and validation confidence, effectively balancing performance and interpretability. 
The framework validates rule effectiveness through the AV identification downstream task---demonstrating strong generalization capability.

Despite these achievements, the current framework has certain limitations. 
The primary limitation concerns velocity computation accuracy. 
Since trajectories are extracted from roadside cameras with oblique viewing angles rather than overhead perspectives, the calculated velocities represent projections onto the image plane rather than true ground speeds. 
This introduces systematic measurement bias, particularly for vehicles moving at angles relative to the camera view. 
While homography transformation partially addresses this issue through spatial calibration, residual errors remain due to imperfect camera positioning and scene geometry assumptions. 
These velocity measurement limitations may affect the precision of speed-related rules, though our validation process helps mitigate their impact on final rule reliability.

Future work can be pursued in several directions. 
First, data diversity can be enhanced by collecting videos from different roadside camera configurations, geographic regions, and AV manufacturers to improve rule generalization across varied traffic conditions and vehicle types. 
Second, more accurate velocity estimation methods can be explored, such as multi-view fusion from multiple synchronized cameras, GPS-based ground truth validation for calibration refinement, or integration of vehicle-to-infrastructure communication data where available. 
Third, computational efficiency can be improved through lightweight model architectures, rule caching mechanisms, and hierarchical reasoning strategies to enable real-time deployment on edge devices. 
Finally, the framework can be extended to complex interaction scenarios by incorporating game-theoretic models and causal reasoning methods to capture multi-agent decision-making dynamics in mixed traffic environments.

\section*{Acknowledgments}

Xiangyu Li and Zhaomiao Guo were supported by the NSF Grant (2521734) and the UT Good Systems Grand Challenge.

Tianyi Wang and Junfeng Jiao were supported by UT Good Systems Grand Challenge.

\bibliographystyle{IEEEtran}
\bibliography{ref}

\newpage

\section{Biography}

\begin{IEEEbiographynophoto}{Xiangyu Li} is currently a Ph.D. student at the Department of Civil, Architectural, and Environmental Engineering, The University of Texas at Austin. He received the M.S. degree in computer engineering from Northwestern University in 2025, and the M.S. degree in transportation engineering from the University of California, Berkeley, in 2023. He earned the B.E. degree in transportation engineering from Beijing Jiaotong University in 2022. His research interests include autonomous driving, intelligent transportation systems, generative AI, and computer vision.
\end{IEEEbiographynophoto}

\begin{IEEEbiographynophoto}{Tianyi Wang} is currently a Ph.D. student at the Department of Civil, Architectural, and Environmental Engineering, The University of Texas at Austin, TX, USA. 
He received his M.S. degree in mechanical engineering and materials science from Yale University, CT, USA, in 2025. 
He earned his B.E. degree in vehicle engineering from Tongji University, Shanghai, China, in 2024. 
His research interests include intelligent transportation systems and connected and automated vehicles, particularly in decision-making, trajectory-planning and cooperative control. 
\end{IEEEbiographynophoto}

\begin{IEEEbiographynophoto}{Junfeng Jiao} received the Ph.D. degree in urban planning from University of Washington in 2010.
He is currently an Associate Professor in the Community and Regional Planning Program at The University of Texas at Austin. 
He is the founding director of Urban Information Lab, director of Texas Smart Cities, director of UT Ethical AI program, and a founding member of UT Austin's Good Systems Grand Challenge. 
His research focuses on Smart Cities, Urban Informatics, and Ethical/Generative AI. 
\end{IEEEbiographynophoto}

\begin{IEEEbiographynophoto}{Christian Claudel} received the Ph.D. degree in electrical engineering from University of California Berkeley in 2010.
He is currently an Associate Professor in the Department of Civil, Architectural, and Environmental Engineering at The University of Texas at Austin. 
His research interests include control and estimation of distributed parameter systems, cyber-physical systems monitoring, and the use of wireless sensor networks for environmental applications.
\end{IEEEbiographynophoto}

\begin{IEEEbiographynophoto}{Zhaomiao Guo} received his Ph.D. degree in transportation system engineering from the University of California at Davis in 2016. He is currently an Assistant Professor in the Department of Civil, Architectural, and Environmental Engineering at The University of Texas at Austin. Dr. Guo's research centers around resilient and intelligent critical infrastructure systems, with a specific emphasis on the coupled and decentralized transportation and energy systems modeling and computation with emerging technologies.
\end{IEEEbiographynophoto}

\end{document}